\documentclass[conference,a4paper]{IEEEtran}

\usepackage{cite}
\usepackage{amsmath,amssymb,amsfonts}
\usepackage{algorithmic}
\usepackage{tikz}
\usepackage{graphicx}
\usepackage{textcomp}
\usepackage{xcolor}
\usepackage{array}
\usepackage{multirow} 
\usepackage{caption}
\usepackage{comment}
\usepackage{pifont}
\usepackage[percent]{overpic}
\usepackage{url}
\usetikzlibrary{shapes.geometric, arrows}
\tikzstyle{block} = [rectangle, rounded corners, minimum width=3cm, minimum height=1cm, text centered, draw=purple, fill=purple!20]
\tikzstyle{arrow} = [thick,->,>=stealth]
\tikzstyle{residual} = [thick,-,>=stealth,draw=black]
\newcommand{\cmark}{\ding{51}}
\newcommand{\xmark}{\ding{55}}
\def\BibTeX{{\rm B\kern-.05em{\sc i\kern-.025em b}\kern-.08em
    T\kern-.1667em\lower.7ex\hbox{E}\kern-.125emX}}
\begin{document}

\title{DAL-PCQA: Enabling Distortion-Level and Language-Driven Reasoning for Point Cloud Quality Assessment}
 
\author{
\IEEEauthorblockN{
Swarna Chakraborty\IEEEauthorrefmark{1},
Gabriel De Castro Araújo\IEEEauthorrefmark{1},
Syeda Tasmi Faria\IEEEauthorrefmark{1}, \\
Marcelo M. Carvalho\IEEEauthorrefmark{2}, and
Mylene C.Q. Farias\IEEEauthorrefmark{1}
}

\IEEEauthorblockA{\IEEEauthorrefmark{1}
\textit{Department of Computer Science, Texas State University, USA} \\
\{wnp23, gfs34, syedafaria, mylene\}@txstate.edu
}

\IEEEauthorblockA{\IEEEauthorrefmark{2}
\textit{Ingram School of Engineering, Texas State University, USA} \\
mmcarvalho@txstate.edu
}
}
\maketitle
%
%
%

%
\begin{abstract}
Point Cloud Quality Assessment (PCQA) methods typically predict scalar Mean Opinion Scores (MOS), which quantify overall perceptual degradation but do not reveal its causes. In contrast, human observers naturally reason in terms of specific distortions such as blur, color shifts, point density changes, missing regions, and geometric deformations. To close this gap, we introduce DAL-PCQA, a distortion-aware, language-annotated dataset for PCQA. DAL-PCQA augments benchmark point clouds with multi-level distortion severity labels, discrete quality categories, and structured natural language descriptions aligned with human perception. We define a point-cloud-specific distortion taxonomy that covers both photometric and geometric artifacts. Statistical analysis reveals characteristic degradation patterns across distortion types and quality levels. To assess the utility of these annotations, we compare zero-shot and fine-tuned multimodal models for generating perceptual quality descriptions. Experiments show that distortion-aware supervision substantially improves lexical and semantic alignment with ground-truth descriptions. By enabling interpretable, distortion-level reasoning, DAL-PCQA facilitates language-driven, explainable point cloud quality assessment. The dataset is publicly available at \url{https://github.com/swarna96/DAL-PCQA}.
\end{abstract}


\section{Introduction}
\label{sec:intro}
Point Clouds (PCs) are a core representation for 3D visual content and are widely used in autonomous driving, robotics, AR/VR, digital twins, and immersive multimedia systems. In immersive and extended reality systems, they enable free-viewpoint navigation and realistic scene interaction, making perceptual quality critical to user experience.  The high dimensionality of PCs imposes substantial storage and bandwidth requirements, making compression necessary to decrease storage needs, support real-time streaming, and allow efficient transmission of PC data without sacrificing perceptual quality. In addition, PCs are highly vulnerable to distortions introduced during acquisition, compression, transmission, and rendering. Unlike 2D images that suffer mainly from spatial artifacts (e.g., blurring, noise, false contours, ringing, etc.), point clouds also exhibit geometric and structural degradations, such as irregular point density, deformed shapes, missing regions, scattering artifacts, and voxelization-induced grid artifacts. These distortions can significantly reduce perceived realism and scene immersion and impact downstream applications. Accurate Point Cloud Quality Assessment (PCQA) is therefore crucial for optimizing compression, benchmarking reconstruction
algorithms, and ensuring good user experience.

A major limitation of traditional PCQA methods is that they usually output only a single scalar value, typically a predicted Mean Opinion Score (MOS). MOS are obtained through subjective experiments in which human participants rate the quality or impairment of a set of stimuli, following standardized recommendations on the setup and procedure of the experiment~\cite{ITU-P910, ITU-R500}, with specific adaptations and new methods developed for PCQA~\cite{su2019perceptual, perry2020quality}. However, while MOS reflects the overall perceived quality of a point cloud, it does not explain \emph{why} a point cloud is judged high- or low quality. Human observers naturally refer to specific distortions, such as whether texture is identifiable, geometry is deformed, or regions are missing. This mismatch between scalar prediction and human-like reasoning hinders interpretability, explainability, and alignment with perceptual semantics.

To address these limitations, we introduce DAL-PCQA\cite{dalpcqa_github}, a structured, distortion-aware, language-annotated dataset for Point Cloud Quality Assessment. Built on SJTU-PCQA\cite{sjtu-pcqa} and WPC\cite{wpc}, it augments point clouds with multi-level annotations over multiple 3D-specific distortion types. Each sample has five-level severity labels per distortion, a discrete quality label, and a natural language description reflecting human perceptual reasoning. By explicitly modeling distortion taxonomy and linguistic descriptions, our framework enables language-driven, explainable PCQA. Distortion-aware descriptions not only add textual output, but also (1) improve interpretability by revealing which photometric and geometric artifacts dominate perceptual degradation, and (2) provide structured supervision that enhances the learned quality representation.

We further demonstrate the utility of the proposed dataset by fine-tuning a language-based quality assessment model and general purpose vision language models (VLM) on point cloud projections and comparing its performance against zero-shot inference. Experimental results show that distortion-aware supervision improves alignment with human perceptual judgments. Although the current dataset is constructed from two static PCQA benchmarks, the proposed annotation framework is dataset-agnostic and can be extended to other static and dynamic point cloud datasets, paving the way for scalable language-driven 3D quality modeling.

The primary contributions of our work are:
\begin{itemize}
    \item We introduce the first structured, distortion-aware language-annotated dataset for point cloud quality assessment, incorporating multi-level 3D distortion labels, discrete quality categories, and human-like descriptions.
    \item We formalize a perceptual distortion taxonomy tailored to geometric and structural artifacts unique to point clouds, bridging the gap between numerical MOS prediction and human reasoning.
    \item We establish a language-driven and explainable PCQA paradigm that goes beyond scalar quality prediction toward an interpretable distortion-aware assessment.
    \item We validate the effectiveness and scalability of the proposed dataset through fine-tuning experiments, demonstrating improved perceptual alignment and extensibility to other PCQA datasets.
\end{itemize}
To facilitate reproducible research, the DAL-PCQA dataset, annotation protocol, and test scripts are available in the DAL-PCQA document github repository\cite{dalpcqa_github}.
\section{Related Works}
Early PCQA work focused on full-reference (FR) methods that require pristine and distorted point clouds~\cite{alexiou2020towards, diniz2020multi, diniz2021color, meynet2020pcqm, Yang2020}. Although effective, FR methods are impractical when no reference is available, motivating No-Reference (NR) PCQA methods~\cite{chetouani2021deep, fan2022no}, which estimate perceptual quality directly from distorted inputs\cite{porcu2024no}. Recent NR approaches increasingly adopt deep learning~\cite{zhu24, liang2024, bourbia2024blind, mithila2025ms, hamidi2025mvaw}, including graph neural networks that model structural relationships within point clouds~\cite{chakraborty2025no, Tilba23, chen2024no}.

Beyond purely geometric modeling, recent multimodal approaches leverage the complementarity between 3D geometry and its 2D projections~\cite{ijcaimm-pcqa, chakraborty2025mt}. Currently, language-guided paradigms are reshaping PCQA. LMM-PCQA~\cite{zhang2024lmm} exploits large multimodal models by converting numerical quality annotations into qualitative natural-language prompts that are spatially aligned with projected views. Pit-QMM~\cite{gupta2025pit} employs LLM-driven cross-modal learning, in which generated textual descriptions are used to supervise and refine the alignment of the cross-modal representations. The FR PCQA framework proposed by Watanabe \textit{et al.}~\cite{watanabe2025full} utilizes multimodal large language models to perform quality reasoning jointly on reference and distorted point clouds. Xie \textit{et al.}~\cite{xie24} introduce an LLM-guided graph-based architecture that incorporates textual supervision to enhance the expressiveness of quality-related representations. Collectively, these studies demonstrate that language models can encode high-level perceptual semantics that are complementary to conventional geometric and visual feature descriptors.

Despite recent progress, language-guided PCQA methods mostly rely on textual supervision from numerical labels or automatically generated prompts and lack a structured, distortion-aware language dataset tailored to point cloud quality reasoning. Language-driven image quality models such as DepictQA\cite{you2024depicting}, Q-Bench\cite{qbench}, and Q-Instruct\cite{wu2024qinstruct} show that large multimodal models can assess visual quality via structured, distortion-aware descriptions aligned with human perception. However, the distortion space of the 3D data fundamentally differs from that of the 2D images. These models are built around 2D pixel-level artifacts (e.g., blur, noise, ringing, false contours) and do not capture geometric degradations unique to point clouds. As a result, directly transferring image-centric language frameworks to 3D data overlooks structural distortions crucial to PCQA.

\section{Proposed Distortion-Aware Language-Annotated PCQA Dataset}
\begin{figure*}[tb]
    \centering
    
    \includegraphics[width=0.95\linewidth]{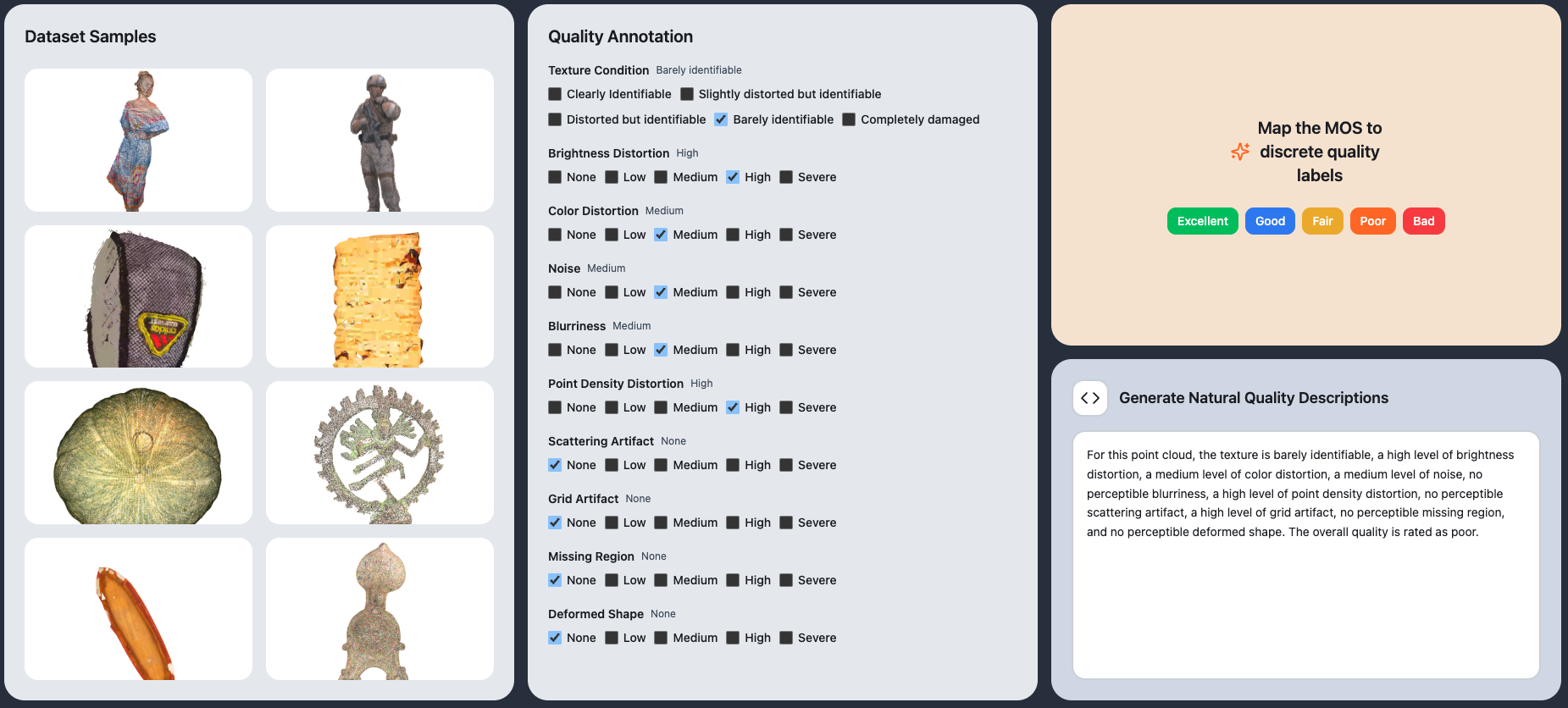}
    \caption{Overview of the DAL-PCQA dataset construction process. Each point cloud is annotated with multi-level distortion severity labels (e.g., texture condition, brightness, color, density, geometric artifacts), mapped from MOS to discrete quality categories, and converted into structured natural language descriptions to enable distortion-aware and language-driven quality assessment.}

    \label{fig:annotationProcess}
\end{figure*}


Figure~\ref{fig:annotationProcess} provides an overview of the DAL-PCQA construction pipeline, encompassing distortion-severity assessment, MOS-to-label mapping, and template-based generation of natural-language quality descriptors. This section details the underlying distortion taxonomy, the annotation protocol, the resulting dataset statistics, and the scalability characteristics of the proposed annotation framework.

\subsection{Distortion Taxonomy}

We propose a point cloud–specific distortion taxonomy for artifacts commonly seen in point clouds. It covers texture condition, brightness distortion, color distortion, noisiness, blurriness, point density distortion, scattering artifacts, grid artifacts, missing regions, and deformed shapes. All categories except texture condition are annotated with five severity levels: None, Low, Medium, High, and Severe. The texture condition is rated by perceptual identifiability, from clearly identifiable to completely damaged. Beyond distortions known from 2D images (brightness, color, noise), this taxonomy explicitly accounts for 3D-specific geometric and structural degradations, such as irregular point density, grid artifacts, missing regions, and deformed shapes that alter the object’s spatial structure and thus require reasoning beyond the pixel-level appearance.

\begin{figure}[tbh]
    \centering
    \includegraphics[width=0.4\linewidth]{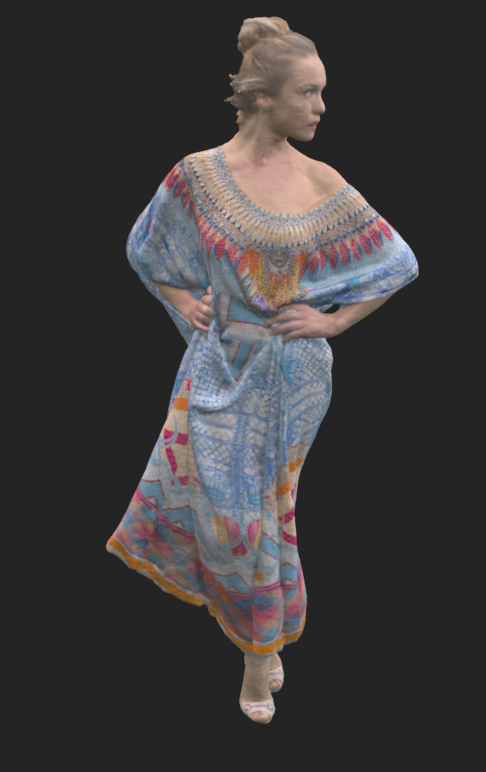}
    \includegraphics[width=0.4\linewidth]{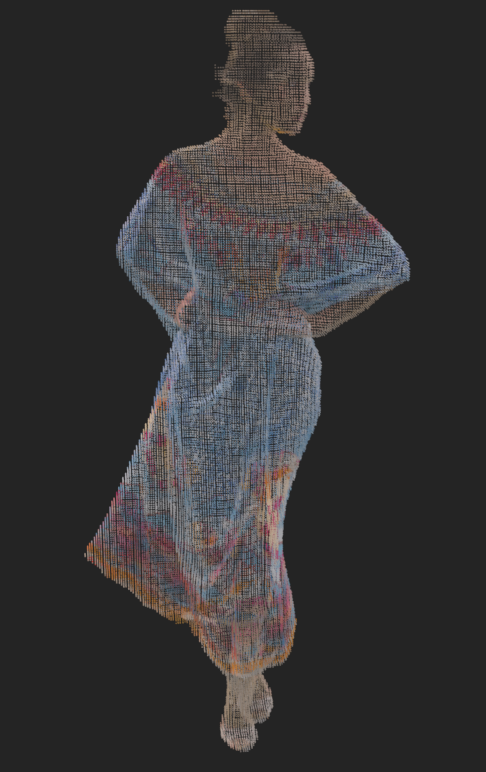}
    \caption{Visual comparison between a reference point cloud (left) and a distorted sample (right). The distorted sample exhibits grid-like sampling artifacts, non-uniform point density, brightness distortion, and structural degradations that are difficult to fully capture using scalar quality scores alone.} \vspace{-0.5cm}
    \label{fig:distortion_example}
\end{figure}

Figure~\ref{fig:distortion_example} illustrates a representative source point cloud from the SJTU-PCQA database, shown both in its original, pristine reference form and in a version affected by distortion. The reference has a uniform distribution of points, smooth surfaces, and a coherent structure. The distorted sample, by contrast, exhibits pronounced grid-like sampling artifacts, spatially non-uniform point density, and the associated degradation of surface smoothness and geometric continuity, as well as observable texture deterioration and luminance/chrominance distortions. In contrast to two-dimensional artifacts, such as blur or compression-induced noise, these distortions modify the intrinsic geometry of the object and are fundamentally three-dimensional in nature. To demonstrate the manner in which the taxonomy encompasses perceptual reasoning, the distorted sample presented in Figure~\ref{fig:distortion_example} is annotated as follows:
\begin{itemize}
    \item \textbf{Texture Condition:} Barely Identifiable
    \item \textbf{Brightness/ Color/ Noise:} High / Medium / Medium
    \item \textbf{Blurriness:} None
    \item \textbf{Point Density Distortion:} High
    \item \textbf{Grid Artifact:} High
    \item \textbf{Other Distortions (Scattering, Missing Region, Deformed Shape):} None
    \item \textbf{MOS:} 3.40625 (Rounded: 3)
    \item \textbf{Label:} Poor
\end{itemize}

The MOS value (3.40625) of the distorted version in Figure \ref{fig:distortion_example} quantifies the overall perceptual degradation; however, it does not provide information regarding which specific distortion types predominantly govern the perceived quality. In contrast, the structured annotation reveals that strong grid artifacts and point density distortion mainly cause the low quality rating. This example underscores the limits of scalar MOS and motivates distortion-aware language-aligned representations that better reflect human perceptual reasoning.

\subsection{Annotation Procedure and Structure}

Distortion annotations were collected through a structured process with three trained annotators. For each point cloud sample, two annotators independently rated all distortion categories to ensure reliability and reduce bias. Annotators followed a detailed protocol defining distortion categories, severity levels, and examples, ensuring consistent interpretation. For each distortion category, the final severity was computed by averaging annotator ratings and rounding to the nearest discrete level, improving robustness, inter-annotator consistency, and perceptual validity.


Building on distortion taxonomy, each dataset entry is designed to preserve perceptual detail and linguistic alignment. Each annotated sample includes the point cloud ID (PLY name), distortion severity for each category, original MOS from the benchmark, rounded MOS, discrete quality label, and structured natural language quality description. Original MOS values are retained as subjective perceptual ground truth and are also assigned five discrete quality labels: \emph{bad}, \emph{poor}, \emph{fair}, \emph{good}, and \emph{excellent}. Since SJTU-PCQA and WPC use different MOS ranges, the mapping is performed according to each dataset's native numerical scale. For SJTU-PCQA, which uses a 1--10 MOS scale, the intervals are 1--2 (\emph{bad}), 3--4 (\emph{poor}), 5--6 (\emph{fair}), 7--8 (\emph{good}) and 9--10 (\emph{excellent}). For WPC, which uses a 1--100 MOS scale, MOS values are normalized and mapped to the five quality levels. This process yields 84 excellent, 271 good, 286 fair, 288 poor, and 188 bad samples.


We use predefined text templates to tightly align structured distortion annotations with natural language supervision while reducing manual effort. Each template refers to the same distortion taxonomy but varies in phrasing. For each point cloud sample, we randomly select a template and automatically fill in the distortion types and severity levels. This guarantees consistent coverage of all distortion categories, adds controlled linguistic variation, and, unlike free-form captions, avoids semantic drift while preserving a direct mapping between annotations and textual descriptions.

\subsection{Dataset Statistics}
Table~\ref{tab:dataset_stats} summarizes the composition of the dataset. The dataset is constructed from two widely used PCQA benchmarks, SJTU-PCQA and WPC, resulting in a total of 1,118 annotated samples. The dataset contains both human-body-centric and object-centric point clouds, ensuring content diversity across semantic categories. Each sample is associated with structured distortion annotations and a corresponding natural language quality description. The label distribution shows that the dataset covers a broad quality spectrum across both source datasets. Although most of the categories are reasonably represented, the \emph{excellent} class is less frequent. This is expected because PCQA benchmarks intentionally introduce distortions at different severity levels, so most samples exhibit some degree of perceptual degradation.


Table~\ref{tab:dataset_comparison} presents a comparative analysis of the proposed dataset against existing PCQA benchmarks. Unlike prior datasets that primarily provide MOS, DAL-PCQA includes explicit distortion-level annotations and curated natural language descriptions, supporting language-guided and interpretable PCQA.

\begin{figure}[tb]
    \centering
    
    \includegraphics[width=0.98\linewidth]{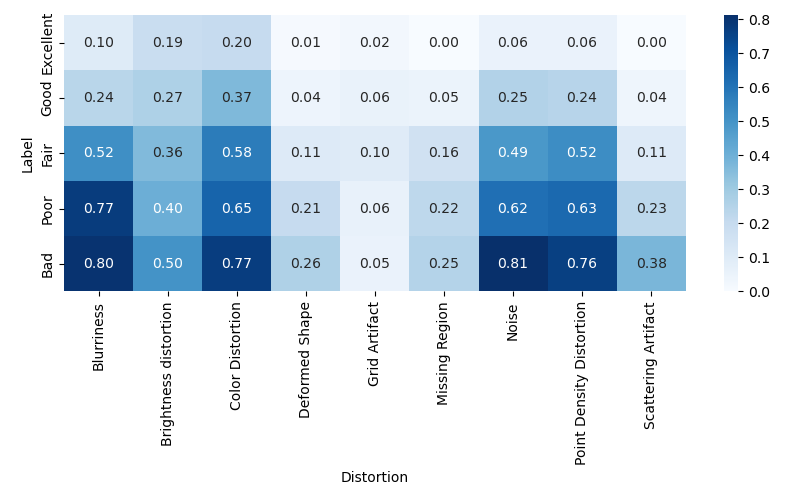}
    \caption{Empirical distortion distribution in the proposed dataset. Lower perceptual quality levels correlate with increased presence of both photometric distortions (blur, noise, color) and geometric artifacts (point density irregularities, deformed shape, missing region), reflecting characteristic degradation patterns in 3D point cloud data.}
    \label{fig:heatmap} \vspace{-0.7cm}
\end{figure}

Figure~\ref{fig:heatmap} shows, for each overall quality label, the percentage of samples whose distortion severity is at least \emph{medium} for each distortion category (columns). The row labels correspond to the overall quality category assigned to each point cloud, while each column represents a specific distortion type. As quality decreases, several distortion categories exhibit a clear and increasing trend. In particular, blur, color distortion, noise, and point density distortion become significantly more prevalent at lower quality levels. Structural distortions, such as missing regions and deformed shapes, are relatively rare in high-quality samples, but become more prominent in \emph{poor} and \emph{bad} quality point clouds. In contrast, grid artifacts, which arise from voxelization or compression-induced regular patterns, remain less frequent across all quality levels.

\begin{table*}[t]
\centering
\caption{Statistical summary of the proposed distortion-aware language PCQA dataset.}
\label{tab:dataset_stats}
\begin{tabular}{lcccccc}
\hline
Source & \#Reference & \#Distorted Samples  & Content Type & Label Distribution \\
 &  &   &  & (Excellent/Good/Fair/Poor/Bad)\\
\hline
SJTU-PCQA & 9 & 378 & Human bodies & 27/124/88/62/76 \\
WPC & 20 & 740 & Objects & 57/147/198/226/112\\
\hline
Overall & 29 & 1118 & Human bodies and objects & 84/271/286/288/188\\
\hline
\end{tabular}
\end{table*}

\begin{table*}[t]
\centering
\caption{Comparison between existing PCQA datasets and the proposed distortion-level language PCQA dataset.}
\label{tab:dataset_comparison}
\begin{tabular}{lcccccc}
\hline
Dataset & \#References & \#Samples & Content Type & MOS & Distortion Annotations \\
\hline
SJTU-PCQA\cite{sjtu-pcqa} & 9 & 378 & Human bodies, Objects & \cmark & \xmark \\
WPC\cite{wpc} & 20 & 740 & Objects & \cmark & \xmark \\
ICIP2020\cite{perry2020quality} & 6 & 90 & Human bodies & \cmark & \xmark \\
PointXR\cite{alexiou2020pointxr} & 5 & 40 & Objects & \cmark & \xmark \\
IRPC\cite{javaheri2020point} & 6 & 54 & Objects & \cmark & \xmark \\
LS-PCQA\cite{liu2023point} & 104 & 22568 & Human bodies, objects, animals & \cmark & \xmark \\

Basics\cite{ak2024basics} & 75 & 1494 & Human bodies, objects, animals, scenes & \cmark & \xmark \\
\hline
Proposed Dataset & 29 & 1118 & Human bodies, objects & \cmark & \cmark \\
\hline
\end{tabular}
\end{table*}

\subsection{Design Rationale and Scalability}

The design of the dataset is guided by three primary objectives. First, it enables distortion-aware learning by preserving fine-grained perceptual information that would otherwise be compressed into a single MOS value. Second, it supports language-driven PCQA frameworks by providing explicit distortion-level annotations and corresponding textual descriptions that align geometric artifacts with textual reasoning. Third, it improves interpretability by allowing model predictions to be analyzed at the distortion level rather than solely through scalar scores.

For the source material, we selected two widely adopted PCQA benchmark datasets that encompass both human-centric and object-centric content. SJTU-PCQA comprises point clouds of human-body exhibiting color and geometric perturbations, including noise, down-sampling, and octree-based compression artifacts. WPC contains a diverse set of object point clouds degraded by down-sampling, Gaussian geometric noise, and G-PCC/V-PCC compression. Collectively, these datasets cover a broad spectrum of acquisition- and compression-induced distortions across heterogeneous content, thereby defining a distortion space that more faithfully represents practical point cloud applications.

Although the current dataset is constructed using two benchmarks, the proposed taxonomy and annotation protocol are dataset-agnostic. The same structured distortion modeling and language alignment process can be applied to other static or dynamic PCQA datasets, enabling the scalable construction of large-scale language-annotated 3D quality corpora. This extensibility positions the dataset not merely as a benchmark augmentation, but as a foundational resource for future explainable and multimodal 3D quality assessment research.

\section{Evaluation of the Proposed Dataset}

\begin{table*}[tb]
\centering
\caption{Comparison of zero-shot and DAL-PCQA fine-tuned settings for multiple MLLMs on WPC and SJTU-PCQA datasets. Best results are indicated by \textbf{bold}.}
\label{tab:text_metrics_all}
\begin{tabular}{llcccccccc}
\hline
Dataset & Model & Setting & BLEU & ROUGE-1 & ROUGE-2 & BERTScore & PLCC & SRCC & LLaMa Score\\
\hline

\multirow{6}{*}{WPC\cite{wpc}}
& \multirow{2}{*}{DepictQA\cite{you2024depicting}}
& Zero-shot  & 0.1094 & 0.2068 & 0.1295 & 0.8433 & 0.1441 & 0.1923 & 2.541\\
& 
& Fine-tuned & \textbf{0.4896} & \textbf{0.7344} & \textbf{0.6051} & \textbf{0.9436} & \textbf{0.6903} & \textbf{0.6946} & \textbf{3.816}\\

& \multirow{2}{*}{LLaVA-1.5-7b\cite{liu2024improved}}
& Zero-shot  & 0.0700 & 0.3155 & 0.1257 & 0.8587 & 0.1118 & 0.1416 & 2.047\\
& 
& Fine-tuned & \textbf{0.5771} & \textbf{0.7670} & \textbf{0.6658} & \textbf{0.9580} & \textbf{0.5953} & \textbf{0.5525}  & \textbf{3.427}\\

& \multirow{2}{*}{InternVL-3-8b\cite{zhu2025internvl3}}
& Zero-shot  & 0.0512 & 0.2863 & 0.1398 & 0.8583 & 0.1912 & 0.1617 & 3.185\\
& 
& Fine-tuned & \textbf{0.5825} & \textbf{0.7975} & \textbf{0.6658} & \textbf{0.9584} & \textbf{0.5469} & \textbf{0.5198} & \textbf{3.472}\\

\hline

\multirow{6}{*}{SJTU-PCQA\cite{sjtu-pcqa}}
& \multirow{2}{*}{DepictQA\cite{you2024depicting}}
& Zero-shot  & 0.1300 & 0.2014 & 0.1323 & 0.8441 & 0.2202 & 0.1548 & 2.512\\
& 
& Fine-tuned & \textbf{0.6865} & \textbf{0.7483} & \textbf{0.7092} & \textbf{0.9642} & \textbf{0.8753} & \textbf{0.8249} & \textbf{4.310}\\

& \multirow{2}{*}{LLaVA-1.5-7b\cite{liu2024improved}}
& Zero-shot  & 0.0217 & 0.2948 & 0.0488 & 0.8570 & 0.1485 & 0.2587 & 1.869\\
& 
& Fine-tuned & \textbf{0.6811} & \textbf{0.8359} & \textbf{0.7391} & \textbf{0.9673} & \textbf{0.8589} & \textbf{0.7838} & \textbf{3.839}\\

& \multirow{2}{*}{InternVL-3-8b\cite{zhu2025internvl3}}
& Zero-shot  & 0.0700 & 0.1757 & 0.1251 & 0.8236 & 0.1728 & 0.1556 & 3.067\\
& 
& Fine-tuned & \textbf{0.7482} & \textbf{0.8270} & \textbf{0.7863} & \textbf{0.9877} & \textbf{0.7702} & \textbf{0.7168} & \textbf{3.618}\\

\hline
\end{tabular}
\end{table*}
\subsection{Experimental Setup}
The primary objective of our evaluation is to determine whether distortion-aware language annotations improve reasoning about 3D quality compared to generic pretrained models. Experiments were conducted on the proposed distortion-aware language-annotated dataset constructed from the SJTU-PCQA~\cite{sjtu-pcqa} and WPC~\cite{wpc} benchmarks. Original MOS values are retained to preserve the perceptual ground truth, whereas distortion annotations and linguistic descriptions serve as supervisory signals for multimodal quality reasoning. To ensure fair evaluation and prevent data leakage, train and test splits are constructed such that there is no overlap of reference content between splits. Specifically, point clouds derived from the same reference model are assigned exclusively to either the training or testing fold.


Currently, most vision-language models (VLMs) are designed to process 2D images rather than 3D point clouds. Although emerging 3D-native models exist, they are not yet adopted for quality reasoning and often require specialized architectural modifications. To ensure compatibility with widely available and reproducible multimodal models, we adopt a projection-based approach in which each point cloud is rendered into four 2D projection images following the procedure in MM-PCQA~\cite{ijcaimm-pcqa}. These  projections serve as visual input to the models during both training and evaluation. We evaluated three representative VLMs: DepictQA~\cite{you2024depicting}, LLaVA~\cite{liu2024improved} and InternVL~\cite{zhu2025internvl3}. DepictQA is specifically designed for image quality reasoning, whereas LLaVA and InternVL are general-purpose VLM. This selection allows us to assess whether the proposed distortion-aware annotations improve performance both for a task-specific quality reasoning model and for general multimodal reasoning models. 

For each model, two settings are considered: 
(1) Zero-shot inference, where the pretrained model is directly applied to the rendered point cloud projections without any additional training on our dataset. 
(2) Fine-tuning, where the model parameters are adapted using the proposed distortion-aware annotations, allowing the model to learn distortion-specific reasoning for point cloud quality. Parameter-efficient fine-tuning is performed using Low-Rank Adaptation (LoRA)~\cite{hu2022lora} with rank $r=16$. Training is conducted for 3 epochs with a maximum token length of 512 using the Adam optimizer with initial learning rate $2\times10^{-4}$ and weight decay $10^{-3}$. 
This design allows us to evaluate whether the proposed annotations provide meaningful supervision beyond generic visual-language knowledge and whether they consistently improve both description quality and perceptual correlation with MOS.

To evaluate whether models generate distortion-aware descriptions consistent with structured annotations, we measure textual alignment between generated outputs and ground-truth descriptions using BLEU~\cite{papineni2002bleu}, ROUGE-1/ROUGE-2~\cite{lin2004rouge}, and BERTScore~\cite{zhangbertscore}. 
BLEU measures n-gram precision and lexical overlap, ROUGE metrics assess recall-based structural similarity, and BERTScore assesses semantic alignment using contextual embeddings.
Although BLEU, ROUGE, and BERTScore measure lexical and semantic similarity, they rely on token overlap and embeddings and may not fully capture distortion correctness. Paraphrased yet accurate descriptions can receive lower scores, while superficially similar text may overlook key distortion details. To complement these metrics with higher-level reasoning, we employ LLaMA~3.1-8b-instruct\cite{grattafiori2024llama} as an LLM-as-a-Judge. Given a generated description and its ground-truth counterpart, the judge model outputs a single integer score from 1 (poor alignment) to 5 (excellent alignment) reflecting overall distortion consistency. We report the  LLaMA Score, computed as the average of these per-sample ratings.

In addition, we assess the extent to which the predicted 5-level quality labels reflect subjective opinion scores by computing the Pearson Linear Correlation Coefficient (PLCC) and the Spearman Rank Correlation Coefficient (SRCC), which characterize, respectively, linear and rank-order agreement with human Mean Opinion Scores (MOS). For this correlation analysis, the predicted 5-level quality categories (Excellent–Bad) are first mapped to ordinal scores on a 1–5 scale, and the correlation is then computed between these ordinal predictions and the corresponding continuous MOS values across all test samples.

\subsection{Impact of Distortion-Aware Supervision}

Table~\ref{tab:text_metrics_all} presents a comparative analysis of the zero-shot and DAL-PCQA fine-tuned configurations of DepictQA, LLaVA, and InternVL on the WPC and SJTU-PCQA datasets. For all three models and across both datasets, fine-tuning with distortion-aware annotations consistently yields marked gains in both language–quality alignment and perceptual correlation metrics. Within the WPC dataset, DepictQA exhibits a substantial improvement in BLEU score, increasing from 0.1094 to 0.4896, and in ROUGE-2, rising from 0.2068 to 0.7344 following fine-tuning. Comparable performance gains are observed for LLaVA (BLEU: 0.0700 → 0.5771) and InternVL (BLEU: 0.0512 → 0.5825), collectively indicating that distortion-aware supervision markedly improves the quality of structured description generation across diverse MLLM architectures. Moreover, BERTScore increases consistently for all evaluated models, further suggesting enhanced semantic alignment with the reference (ground-truth) descriptions.

The correlation between predicted quality labels and subjective MOS also improves markedly. For example, on WPC, DepictQA’s PLCC increases from 0.1441 to 0.6903 and SRCC from 0.1923 to 0.6946. LLaVA and InternVL similarly exhibit substantial gains in PLCC and SRCC after fine-tuning, confirming that distortion-aware supervision improves not only textual fidelity, but also perceptual consistency with human judgments. A similar trend is observed on SJTU-PCQA. Zero-shot MLLMs produce relatively weak lexical alignment and low correlation with MOS, whereas fine-tuning significantly improves both language metrics and correlation scores across all models. In particular, fine-tuned InternVL and LLaVA achieve strong BERTScore and correlation values, suggesting that the proposed distortion-aware annotations provide model-agnostic supervision benefits. The LLaMA Score further confirms this trend by showing consistent increases in the mean alignment score across all models and datasets, demonstrating that distortion-aware supervision enhances high-level semantic consistency beyond token-level similarity metrics.

Overall, these results demonstrate that DAL-PCQA enables an effective adaptation of MLLMs to 3D distortion reasoning. The improvement is mainly driven by structured supervision that links projected point cloud views with distortion categories, severity levels, and quality labels. In contrast to a zero-shot setting, MLLMs rely on generic image-language priors and often produce broad appearance-level descriptions. Fine-tuning exposes the models to explicit associations among projected point cloud views, distortion categories, severity levels, and quality labels, enabling more distortion-specific reasoning. This explains the consistent gains in both textual alignment and MOS correlation across architectures. However, gains may be smaller for subtle artifacts, rare distortion types, or geometry-dependent degradations that are not fully captured by projection-based inputs.


\section{Conclusion}
In this paper, we introduced a structured, distortion-aware, language-annotated dataset for point cloud quality assessment. Unlike conventional PCQA datasets that primarily provide scalar MOS values, the proposed dataset integrates multi-level distortion severity annotations, discrete quality labels, and structured natural language descriptions aligned with human perceptual reasoning. By explicitly modeling both photometric and geometric degradations, the dataset bridges the gap between numerical quality prediction and interpretable distortion-level assessment.

Statistical analysis shows that the dataset captures realistic multi-distortion patterns across quality levels, while experiments demonstrate that distortion-aware fine-tuning improves textual alignment and MOS correlation compared with zero-shot inference.

Future work will extend the annotation protocol to additional static and dynamic datasets and explore native 3D multimodal architectures that directly process raw point clouds without projection. In addition, while the current template-based language generation ensures distortion consistency, we plan to incorporate free-form human-authored captions to increase linguistic diversity and further enhance natural perceptual reasoning.


\bibliographystyle{ieeetr}
\bibliography{strings,refs}

\end{document}